# Unifying Topic, Sentiment & Preference
# in an HDP-Based Rating Regression Model for Online Reviews

Zheng Chen[1], Yong Zhang[1,2], Yue Shang[1], Xiaohua Hu[1]

[1] Drexel University, USA

[2] China Central Normal University, China

**Abstract**

This paper proposes a new HDP based online review rating regression model named Topic-Sentiment-Preference Regression Analysis (TSPRA). TSPRA combines topics (i.e. product aspects), word sentiment and user preference as regression factors, and is able to perform topic clustering, review rating prediction, sentiment analysis and what we invent as "critical aspect" analysis altogether in one framework. TSPRA extends sentiment approaches by integrating the key concept "user preference" in collaborative filtering (CF) models into consideration, while it is distinct from current CF models by decoupling "user preference" and "sentiment" as independent factors. Our experiments conducted on 22 Amazon datasets show overwhelming better performance in rating predication against a state-of-art model FLAME (2015) in terms of error, Pearson's Correlation and number of inverted pairs. For sentiment analysis, we compare the derived word sentiments against a public sentiment resource SenticNet3 and our sentiment estimations clearly make more sense in the context of online reviews. Last, as a result of the de-correlation of "user preference" from "sentiment", TSPRA is able to evaluate a new concept "critical aspects", defined as the product aspects seriously concerned by users but negatively commented in reviews. Improvement to such "critical aspects" could be most effective to enhance user experience.

## 1 Introduction

The rapidly growing amount of rated online reviews brings great challenges as well as needs for regression analysis on review ratings. It has been studied in various publications that factors like review texts, product aspects, word sentiments, aspect sentiments, user clusters, and product clusters, have been integrated in the regression analysis. Models built from such analysis of course can be applied to "predict" ratings of unannotated reviews, meanwhile the factor estimations obtained from regression may provide insight about users or products and is helpful for decision makers to conduct business activities.

Many recent review models [1], [2], [3], etc. adopt a topic modeling method Latent Dirichlet Allocation (LDA) [4] as the modeling framework. In LDA, product aspects involved in an online review are treated as topics. In future discussion, "topics" and "product aspects" are synonyms and equivalent.

A further development of topic modeling methods is Hierarchical Dirichlet Process (HDP) [5], which replaces the Dirichlet allocation in LDA with Dirichlet processes. The distinct advantage of employing the Dirichlet process is that the number of topics is inferred by the model from data rather than specified. As a result, it becomes unnecessary to experiment on aspect numbers as is done in [1]. Experiments to compare these two models have consistently shown HDP transcends LDA in terms of perplexity in different settings, see [6] and [7]. However, all review models examined in this paper are not based on the HDP.

In this paper we build a new review model *Topic-Sentiment-Preference Regression Analysis* (TSPRA) based on the HDP framework. In TSPRA, topics, sentiments and user preferences are considered as regression factors. TSPRA is driven by three major considerations. The first one is

2to develop an automatic approach to evaluate word sentiments. Until recently, many sentiment analyses still depend on manually annotated corpus for model training. To circumvent laborious annotating efforts and the potential danger that artificial annotations are done by uninterested persons, exploitation of online reviews and their ratings, which are naturally annotated corpus, has become increasingly popular among researchers to conduct sentiment analysis. In this paper we develop a new rating regression model to compute word sentiments automatically based on online reviews and their ratings.

The second consideration is to experiment on a new idea of redefining the concept "user preferences" that have been addressed by previous collaborative filtering models, such as those discussed in section 2. There is no doubt user preference is a crucial and necessary factor in rating regression. For example, suppose we have a customer in a restaurant that cares more about taste rather than price. If the taste of the restaurant does not meet his expectation, most likely he would not give a good rating, no matter how good the price is. Otherwise he may give a positive rating even if the dinner is a little expensive. However, after careful examination of recent CF papers [1] [8] [9] [10] that include user preference in the modeling, we found the term is not clearly defined and it functions more like topic-level sentiment in the models.

We propose that it might be necessary to distinguish between user preference and sentiment. Intuitively, in the restaurant example, one's preference in taste does not necessarily mean **one** must give the restaurant's taste a high score. However, in CF models mentioned in section 2, user preferences act more like aspect-level sentiments because it is typically modelled as the sole type of factors that directly result in the overall ratings. For example, [8] models both word sentiments and review ratings as consequence of aspect "interests", therefore word sentiments exert effect on review ratings through the "interests" and must have a strong correlation with the "interests". Likewise in [9] the aspect scores are designed to cause ratings and sentence sentiments, thus the sentence sentiments must have a strong correlation with the aspect scores. Such correlation contradicts with our intuition and in model design it might also reduce the amount of information that can be unfolded through statistical inference. Our model TSPRA de-correlates user preference from sentiment by explicitly defining it as "how much a customer cares about a product aspect" and it is independent from sentiments. The difference is illustrated in Fig 1 (a) and (b).

This change has a significant impact – now a high/low preference is allowed to co-occur with negative words and a low/high rating. This makes sense in real application. If one cares a lot about cellphone battery and buys a cellphone with a poor battery, one could give a negative comment with many negative words that lead to a low rating. By contrast, in CF models, a high "preference" typically implies a high rating and positive words. This change will be verified by a weak Pearson's correlation value as discussed in section 4.4.

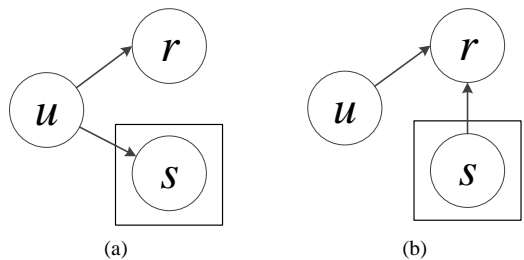

Fig 1 (a) user preference $u$ in two CF models [8] and [9] is designed to cause both the review rating $r$ and word/sentence sentiments $s$, leading to a strong correlation between $u$ and $s$; (b) in our model preference $u$ and sentiment $s$ are designed as independent variables that co-determine the review rating.

The third concern lies in the number of product aspects. If we have many heterogeneous datasets, it would be laborious to find out appropriate topic number for each dataset, while experiment results on topic number based on one or two datasets, as done in [1], might be unwarranted for other datasets. A further development of topic model, the HDP [5] where HDP stands for *Hierarchical Dirichlet Process*, help solve this issue. By taking advantage of the Dirichlet stochastic process we no longer need to specify the topic number for each dataset beforehand. Also in various settings it has been proven HDP models transcend LDA models [7] [6] in terms of perplexity, thus adoption of HDP model as the framework of a review model is promising.

The rest of this paper is arranged as follows. Section 2 discusses previous review models and the HDP model. Section 3 introduces the setup of our regression model, the inference and prediction formulas. We demonstrate our experiment results in Section 4 and make conclusions and some remarks on future work in Section 5.

## 2 Related Work

Two main approaches are studied in recent publications regarding online review rating regression: the *collaborative filtering* (CF) approaches and the *sentiment* approaches. Collaborative filtering approaches are active research topics in rating regression and prediction. There are models not exploiting review contents, such as [11] [10], and they typically depend on factorization of the user-product rating matrix. More recent models [12] [13] [14] endeavor to incorporate texts as additional resource beyond ratings to add to their predictive power and recognize product aspects. In these models, each aspect is associated with "preference" scores and the overall review rating is viewed as the aggregation of these aspect scores. Further developments unify sentiments into the CF framework. A recent study [9] associates each aspect with a "rating distribution" and views sentence sentiments as the consequence of these rating distributions. Another recent work [8] call the aspect scores as "interests" and combines word sentiments into their model. Both models are similar to our model in comparison to other CF models mentioned above, however, on one hand the "preferences" or "rating distributions" or "interests" feel somewhat like aspect-level sentiment given that they measure if customers like or dislike a product aspect; on the other hand the main objective of CF models is still doing recommendation, therefore sentiment, especially word sentiment, is a subordinate unit of analysis. In all these CF models, [8] is the only CF model that can infer word sentiments.

The other perspective of rating regression is to consider review ratings as a direct consequence of word sentiments. [15] proposes a probabilistic rating regression model that requires aspect keyword supervision, and the follow-up studies [2] removes the keyword supervision by merging LDA into the model to perform keyword assignment. Both papers can be viewed as a further extension to the work dedicated to combine aspects and sentiments – see [16] [17] [3]. Intuitively a key difference between these "sentiment" models and the CF models with sentiment is that in "sentiment" models the review ratings directly result from the aggregation of word sentiments, while the CF approaches treat word sentiments and review ratings as consequences of per-aspect user preferences.

Many of the above-mentioned works are developed from the topic model framework, typically LDA [4], due to the fact that a reviewer usually comments on different aspects of a target item and the strong analogy between product aspects and topics. A direct extension to LDA is [5] which introduces a stochastic process name hierarchical Dirichlet process to infer number of topics from the data. Further improvements targeting rating regression include [18] and [19]. No matter "preference" or "sentiment", they are more meaningful with respect to a certain topic. User preference is usually regarding a product aspect. A word might be positive in one topic but negative in another topic. These models that adopt the topic model framework assign each topic a score, whether it is "preference", "interest" or "sentiment", and the overall review rating is viewed as an aggregation of the aspect scores.

# 3 Model Construction

## 3.1 Model Specification

Our model consists of four parts: the topic part, the sentiment part, the user preference part and the rating regression part. The topic part is the same as HDP [5], and the rest are our extensions, as shown in Fig 2. These four parts together describe the generative process of observable data: review text and ratings.

We include a brief explanation of HDP (the part in the dashed box of Fig 2) for the sake of completeness. In HDP, $G_0$ and $G_d$ are global and local random probability measures of topics defined by a two-level hierarchical Dirichlet process. This stochastic process produces $K$ topics $\varphi_1, \ldots, \varphi_K$ where the number of topics $K$ is dynamic. Moreover, each word $w_{di}$ in the review $d$ is viewed as being drawn from one of the latent topics indexed by $z_{di}$, where $z_{di} = 1, \ldots, K$. HDP is intractable and it has to be realized through "representations". Specifically in our model we use the Chinese Restaurant Franchise (CRF) representation, which is a two-level Chinese Restaurant Process (CRP). By the CRF representation, each document is metaphorically a restaurant of the same franchise, and each word of the document is viewed as a customer dining at one restaurant. The first customer (the first word) must sit at a new table and order a "dish", where "table" is a metaphor of word cluster, "dish" is a metaphor of topic, and by "order a dish" we mean a topic is associated with the table and hence all words at the table. The next customer (the next word) either chooses an existing table to sit and share the "dish" of the table (i.e. the word is put in an existing cluster and is assigned to the topic associated with that table), or sits at a new table and orders a new dish (i.e. the new word is put into a new cluster and the cluster and a topic is assigned to the new cluster). All documents share the same set of topics, analogous to that all restaurants are under the same franchise serving the same set of dishes. In CRF representation, each $z_{di}$ is determined by a latent table index $t_{di}$ assigned to word $w_{di}$ and a latent topic index $k_{dt}$ assigned to table $t$ in review $d$, i.e. $z_{di} = k_{dt_{di}}$.

For the sentiment part, $\pi_{K \times V}$ are $K \times V$ sentiment distributions where $S$ is the number of sentiment polarities and $V$ is the vocabulary size. In this paper $S$ is set 3 following the convention of sentiment analysis, indicating each word can be positive, negative or neutral. A word $w$ under topic $\varphi_k$ is associated with a three-dimensional sentiment distribution $\pi_{kw}$, such as 0.2, 0.4, 0.4 where the numbers are the probabilities of the word $w$ being positive, negative or neutral under topic $\varphi_k$. We denote the latent sentiment index of word $w_{di}$ as $s_{di}$, then $s_{di}$ is sampled from the sentiment distribution $\pi_{z_{di}w_{di}}$. We let $s_{di} = -1, 0, +1$ represent negative, neutral and positive sentiment respectively.

The preference part is similar to the sentiment part, where $\psi_{K \times X}$ are $K \times X$ preference distributions over $U$ strengths of preferences and $X$ users. In this paper $U$ is set 2 indicating user preference can be strong or weak. Now an author $x$ under topic $\varphi_k$ is associated with a two-dimensional preference distribution $\psi_{kx}$, such as 0.2, 0.8 where the numbers are the probabilities of the author $x$ having strong or weak preference to topic $\varphi_k$. We denote the author of review $d$ by $x_d$, then there is also a preference index $u_{di}$ associated with each word $w_{di}$, which is sampled from $\psi_{z_{di}x_d}$. We let $u_{di} = 0, 1$ represent strong and weak preference respectively.

Rating regression part is the generative process of the review rating $r_d$ of document $d$ (the rating of an online review is typically a 1-5 scale). For each word, when the latent variables $u_{di}$ and $s_{di}$ are given, the rating variable $r_{di}$ is determined by the association rule in Table 1. In the association rule, $\mu$ ($1 \leq \mu \leq 5$) is a parameter of our model named "neutral rating". For a review rating of 1-5 scale, one might think the middle value $\mu = 3$ is the neutral rating. In fact, this might not be true. We will see later in section 4 that in real world users tend to give positive ratings, and the experiment shows a better $\mu$ is larger than 3. We have the following two main reasons for this associations rule.

First, this association rule is intuitive, simple and straightforward. Our assumption is that if a user gives a fiver star rating, one must exhibit high concern as well as positive opinion. Therefor we map $u_{di} = 1, s_{di} = 1$ to $r_{di} = 5$. If that product aspect is less concerned, i.e. $u_{di} = 0$, then the consequent rating should be lower even if the sentiment is still positive. Hence $u_{di} = 0, s_{di} = 1$ is mapped to $r_{di} = \frac{5+\mu}{2}$. Mapping $u_{di} = 1, s_{di} = -1$ to $r_{di} = 1$, and mapping $u_{di} = 0, s_{di} = -1$ to $r_{di} = \frac{1+\mu}{2}$ are for similar reasons. If $s_{di} = 0$, meaning the user has no sentiment orientation, then the rating is the neutral value $\mu$ regardless of preference.

Secondly, the association rule is our current solution to decouple sentiments from user preferences. By the rule, we can see a strong user preference can be accompanied by a negative sentiment, and a weak preference can co-occur with a positive sentiment. This is different from schemes in previous CF models like [8] [9], where user preferences over topics in a review are first sampled according to some distribution (e.g. beta distribution, normal distribution) based on review rating, and word sentiments under a topic are then sampled based on user preferences of that topic. As a result, a high rating will indicate both strong preference and positive sentiment, leading to a correlation between sentiments and preferences, and we argue that in those CF models user preferences function more like topic-level sentiments.

Table 1. The association rules for rating

- If $u_{di} = 1$ (strong preference) and $s_{di} = -1$ (negative sentiment), then $r_{di} = 1$.
- If $u_{di} = 0$ (weak preference) and $s_{di} = -1$, then $r_{di} = \frac{1+\mu}{2}$.
- If $u_{di} = 1$ and $s_{di} = +1$ (positive sentiment), then $r_{di} = 5$.
- If $u_{di} = 0$ and $s_{di} = +1$, then $r_{di} = \frac{5+\mu}{2}$
- Otherwise $r_{di} = \mu$. This rating is named neutral word rating, and a word with neutral word rating is named a neutral word.

Once all word ratings are determined, the review rating $r_d$ is drawn according to $r_d \sim N(\overline{r_{di}}, \sigma^2)$, where $\overline{r_{di}}$ is the mean of non-neutral word ratings (i.e. average of word ratings excluding neutral words), and $\sigma^2$ is the rating noise, and $N(\overline{r_{di}}, \sigma^2)$ is the normal distribution with $\overline{r_{di}}$ as mean and $\sigma^2$ as variance.

The above is the complete setup of the generative process assumed by our model, illustrated by the plate diagram in Fig 2, and we summarize the generative process as the following,

- Define the topic prior $H$ for global topic generation as a $V$-dimensional Dirichlet distribution $\text{Dir}(\beta)$, where $V$ is the vocabulary size. Define $G_0$ as the global random probability measure distributed according to Dirichlet process $\text{DP}(H, \gamma)$ where $\gamma$ is the concentration parameter.
- For each topic $k$
  - Draw a topic distribution over vocabulary $\varphi \sim \text{Dir}(\beta)$
  - For each word $w$ in the vocabulary, draw a distribution over 3-dimensional sentiment $\pi_{kw} \sim \text{Dir}(\lambda)$.
  - For each author $x$, draw a user preference distribution over 2-dimensional strength $\psi_{kx} \sim \text{Beta}(\eta)$.
- For each review $d$ authored by the user $x_d$
  - Draw a local random probability measure $G_d$ from $G_0$ distributed according to Dirichlet process $\text{DP}(G_0, \alpha)$ where $\alpha$ is the concentration parameter.
  - For each $i$th word $w_{di}$
    - Draw a table $t_{di}$ and a topic $z_{di}$ from $G_d$ and $G_0$ according to CRF representation of HDP
    - Draw the word $w_{di} \sim \text{Multinomial}(z_{di})$
    - Draw a sentiment $s_{di} \sim \text{Multinomial}(\pi_{z_{di}w_{di}})$
    - Draw a user preference of author $x_d$ given topic $z_{di}$, $u_{di} \sim \text{Binomial}(\psi_{z_{di}x_d})$
  - Compute a latent variable $r_{di}$ and sample $r_d$ according to the rules as shown in Table 1.

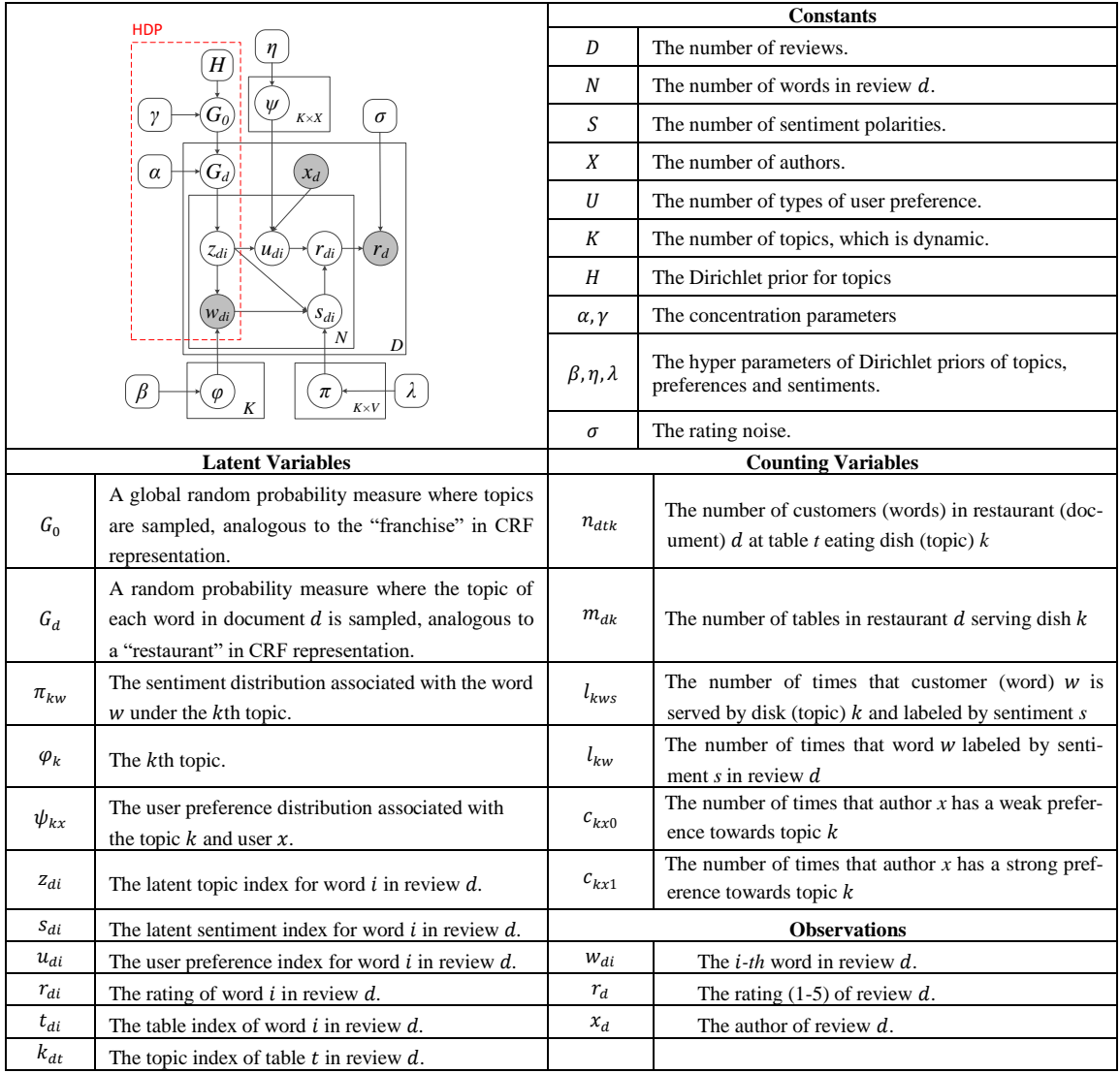

| | Constants |
|---|---|
| $D$ | The number of reviews. |
| $N$ | The number of words in review $d$. |
| $S$ | The number of sentiment polarities. |
| $X$ | The number of authors. |
| $U$ | The number of types of user preference. |
| $K$ | The number of topics, which is dynamic. |
| $H$ | The Dirichlet prior for topics |
| $\alpha, \gamma$ | The concentration parameters |
| $\beta, \eta, \lambda$ | The hyper parameters of Dirichlet priors of topics, preferences and sentiments. |
| $\sigma$ | The rating noise. |

| | Latent Variables | | Counting Variables |
|---|---|---|---|
| $G_0$ | A global random probability measure where topics are sampled, analogous to the "franchise" in CRF representation. | $n_{dtk}$ | The number of customers (words) in restaurant (document) $d$ at table $t$ eating dish (topic) $k$ |
| $G_d$ | A random probability measure where the topic of each word in document $d$ is sampled, analogous to a "restaurant" in CRF representation. | $m_{dk}$ | The number of tables in restaurant $d$ serving dish $k$ |
| $\pi_{kw}$ | The sentiment distribution associated with the word $w$ under the $k$th topic. | $l_{kws}$ | The number of times that customer (word) $w$ is served by disk (topic) $k$ and labeled by sentiment $s$ |
| $\varphi_k$ | The $k$th topic. | $l_{kw}$ | The number of times that word $w$ labeled by sentiment $s$ in review $d$ |
| $\psi_{kx}$ | The user preference distribution associated with the topic $k$ and user $x$. | $c_{kx0}$ | The number of times that author $x$ has a weak preference towards topic $k$ |
| $z_{di}$ | The latent topic index for word $i$ in review $d$. | $c_{kx1}$ | The number of times that author $x$ has a strong preference towards topic $k$ |
| $s_{di}$ | The latent sentiment index for word $i$ in review $d$. | | Observations |
| $u_{di}$ | The user preference index for word $i$ in review $d$. | $w_{di}$ | The $i$-th word in review $d$. |
| $r_{di}$ | The rating of word $i$ in review $d$. | $r_d$ | The rating (1-5) of review $d$. |
| $t_{di}$ | The table index of word $i$ in review $d$. | $x_d$ | The author of review $d$. |
| $k_{dt}$ | The topic index of table $t$ in review $d$. | | |

Fig 2 The graphic presentation of our regression model, Topic-Sentiment-Preference Regression Analysis (TSPRA). Our model is able to perform topic-clustering, word-level sentiment analysis, topic-level preference analysis and rating regression simultaneously.

Comparing to HDP, our model adds three more latent variables for each word $s_{di}$, $u_{di}$ and $r_{di}$, where $r_{di}$ is generated by association rules given $s_{di}$ and $u_{di}$. For counting variables, marginal counts are represented with dots. Thus $n_{dt\cdot}$ represents the number of customers in restaurant $d$ at table $t$, $n_{d\cdot k}$ represents the number of customers in restaurant $d$ eating dish $k$, $m_{d\cdot}$ represents the number of tables in restaurant $d$, $m_{\cdot k}$ represents the number of tables serving dish $k$, and $m_{\cdot\cdot}$ is the total number of tables occupied.

### 3.2 Inference

In training, the variables in our model TSPRA are estimated by a collection of reviews and their ratings. We adopt the Gibbs sampling [5] [20] to obtain latent variable estimation under the CRF representation of HDP and Dirichlet priors.

Based on CRF and the generative process illustrated in Fig 2, we have six sets of latent variables: **z**, **t**, **k**, **s**, **u** and **r** – the topic indexes, the table indexes of all words, the topic indexes of all tables, the word sentiments, the per-topic user preferences and word ratings. Note that $\mathbf{z}$ is determined by $\mathbf{t}$ and $\mathbf{k}$ as discussed in section 3.1, and $\mathbf{r}$ is determined by $\mathbf{s}$ and $\mathbf{u}$ using rules in Table 1, therefore the whole state space are determined by $(\mathbf{t}, \mathbf{k}, \mathbf{r})$. The conditional distributions of this state space are iteratively re-sampled by the Gibbs sampler described below. Provided a sufficient number of re-sample iterations, the final estimation of latent variables can be viewed as a sample generated from the process in Fig 2, for detailed discussion, please refer to [21].

***Re-sample t***. We first re-sample the table index $t_{di}$ of each word $w_{di}$ according to equation (1) where the tilde "∼" denotes "all other variables and parameters".

$$p(t_{di} = t|\mathbf{t}^{-t_{di}}, \sim) = \begin{cases} \frac{n_{dt\cdot}^{-di}}{n_{d\cdot\cdot}^{-di} + \alpha} f_{k_{dt}}^{-w_{di}}(w_{di}, r_d|z_{di} = k_{dt}, \sim) & t \text{ is previously used} \\ \frac{\alpha}{n_{d\cdot\cdot}^{-di} + \alpha} p(w_{di}, r_d|t = t_{new}, \sim) & t = t_{new} \end{cases}$$

$$\propto \begin{cases} n_{dt\cdot}^{-di} f_{k_{dt}}^{-w_{di}}(w_{di}, r_d|z_{di} = k_{dt}, \sim) & t \text{ is previously used} \\ \alpha p(w_{di}, r_d|t = t_{new}, \sim) & t = t_{new} \end{cases} \quad (1)$$

In the situation when a word chooses an existing table, i.e. "$t$ is previously used", the topic of the chosen table is $\varphi_{k_{dt}}$, then the likelihood of observation $w_{di}$ given $z_{di}$ and other variables is estimated using equation (2).

$$f_{k_{dt}}^{-w_{di}}(w_{di}, r_d|z_{di} = k_{dt}, \sim) = \sum_{u_{di}} \sum_{s_{di}} p(w_{di}, r_d, u_{di}, s_{di}|z_{di}, \sim)$$
$$= \sum_{u_{di}} \sum_{s_{di}} p(w_{di}|z_{di}, \sim) p(u_{di}|z_{di}, \sim) p(s_{di}|z_{di}, w_{di}, \sim) p(r_d|\overline{r_{d\iota}}, \sim) \quad (2)$$

In the situation when a customer chooses a new table, a topic needs to be sampled for the new table using equation (3).

$$p(k_{dt^{new}} = k|\sim) = \begin{cases} \frac{m_{\cdot k}}{m_{\cdot\cdot} + \gamma} f_k^{-w_{di}}(w_{di}, r_d|\varphi_k, \sim) & k \text{ is previously used} \\ \frac{\gamma}{m_{\cdot\cdot} + \gamma} f_{k^{new}}^{-w_{di}}(w_{di}, r_d|\varphi_{k^{new}}, \sim) & k = k^{new} \end{cases}$$

$$\propto \begin{cases} m_{\cdot k} f_k^{-w_{di}}(w_{di}, r_d|\varphi_k, \sim) & k \text{ is previously used} \\ \gamma f_{k^{new}}^{-w_{di}}(w_{di}, r_d|\varphi_{k^{new}}, \sim) & k = k^{new} \end{cases} \quad (3)$$

Hence in equation (1) the likelihood of observing $(w_{di}, r_d)$ given the new table, which is denoted by $p(w_{di}, r_d|t_{di} = t_{new}, \sim)$, can be estimated by equation (4).

$$p(w_{di}, r_d|t_{di} = t_{new}, \sim) = \sum_{k=1}^{K} \frac{m_{\cdot k}}{m_{\cdot\cdot} + \gamma} f_k^{-w_{di}}(w_{di}, r_d|\varphi_k, \sim) + \frac{\gamma}{m_{\cdot\cdot} + \gamma} f_{k^{new}}^{-w_{di}}(w_{di}, r_d|\varphi_{k^{new}}, \sim) \quad (4)$$

The rest of our inference for re-sample of table index $t_{di}$ is how to estimate the likelihood $f_k^{-w_{di}}(w_{di}, r_d|\varphi_k, \sim)$. By equation (2) it is not hard to derive equation (5), in which for notation clarity we simplify $w_{di}$ as $w$, $s_{di}$ as $s$, and $u_{di}$ as $u$.

$$f_k^{-w}(w, r_d|\varphi_k, \sim) = f_k^{-w}(w|\varphi_k, \sim) f_k^{-w}(r_d|w, \varphi_k, \sim)$$
$$\propto \frac{l_{kw} + \beta}{l_{k\cdot} + V\beta} \times \sum_u \sum_s \frac{c_{kxu} + \eta}{c_{kx\cdot} + U\eta} \times \frac{l_{kws} + \lambda}{l_{kw\cdot} + S\lambda} \times e^{-\frac{(r_d - \overline{r_{di}})^2}{2\sigma^2}} \quad (5)$$

A special case when $k = k^{new}$ is given by equation (6).

$$f_{k^{new}}^{-w}(w, r_d|\varphi_{k^{new}}, \sim) \propto \frac{1}{V} \times \frac{1}{U} \times \frac{1}{S} \times \sum_u \sum_s e^{-\frac{(r_d - \overline{r_{di}})^2}{2\sigma^2}} \quad (6)$$

***Re-sample k***. We then re-sample the topic index $k_{dt}$ of each table $t$ in review $d$ by equation (7). Note in this step all words associated with table $t$ might switch topic.

$$p(k_{dt} = k|\mathbf{k}^{-k_{dt}}, \sim) = \begin{cases} \frac{m_{\cdot k}}{m_{\cdot\cdot} + \gamma} f_k^{-\mathbf{w}_{dt}}(\mathbf{w}_{dt}, r_d|\varphi_k, \sim) & k \text{ is previously used} \\ \frac{\gamma}{m_{\cdot\cdot} + \gamma} f_{k^{new}}^{-\mathbf{w}_{dt}}(\mathbf{w}_{dt}, r_d|\varphi_{k^{new}}, \sim) & k = k^{new} \end{cases}$$

$$\propto \begin{cases} m_{\cdot k} f_k^{-\mathbf{w}_{dt}}(\mathbf{w}_{dt}, r_d|\varphi_k, \sim) & k \text{ is previously used} \\ \gamma f_{k^{new}}^{-\mathbf{w}_{dt}}(\mathbf{w}_{dt}, r_d|\varphi_{k^{new}}, \sim) & k = k^{new} \end{cases} \quad (7)$$

In the situation when an existing topic is chose for table $t$ in review $d$, the likelihood of observation $(\mathbf{w}_{dt}, r_d)$ given topic $\varphi_k$ is estimated by equation (8). The same notation simplification applied to equation (5) also applies here, i.e. we simplify $\mathbf{w}_{dt}$ as $\mathbf{w}$, $\mathbf{s}_{dt}$ as $\mathbf{s}$, $u_{dt}$ as $u$, and $s_{di}$ as $s$.

$$f_k^{-\mathbf{w}}(\mathbf{w}, r_d | \varphi_k, \sim) = f_k^{-\mathbf{w}}(\mathbf{w}|\varphi_k, \sim) f_k(r_d|\mathbf{w}, \varphi_k, \sim)$$
$$\propto (\prod_{w \in \mathbf{w}} \frac{l_{kw} + \beta}{l_{k \cdot} + V\beta}) \times \sum_u (\frac{c_{kxu} + \eta}{c_{kx \cdot} + U\eta} \sum_\mathbf{s} \prod_{w \in \mathbf{w}} \frac{l_{kws} + \lambda}{l_{kw \cdot} + S\lambda} \times e^{-\frac{(r_d - \overline{r_{di}})^2}{2\sigma^2}}) \quad (8)$$

Equation (8) is way too complicated to be evaluated in full, especially for the part summing over $\mathbf{s}$, which leads to a sum of $S^{|\mathbf{w}|}$ terms. We might instead use the following approximation.

$$f_k^{-\mathbf{w}}(\mathbf{w}, r_d|\varphi_k, \sim) \approx f_k^{-\mathbf{w}}(\mathbf{w}, \mathbf{r}_d|\varphi_k, \sim) \propto (\prod_{w \in \mathbf{w}} \frac{l_{kw} + \beta}{l_{k \cdot} + V\beta}) \times \frac{c_{kxu} + \eta}{c_{kx \cdot} + U\eta} \times \prod_{w \in \mathbf{w}} \frac{l_{kws} + \lambda}{l_{kw \cdot} + S\lambda} \quad (9)$$

A special case when $k = k^{new}$ is given by equation (10).

$$f_{k^{new}}^{-\mathbf{w}}(\mathbf{w}, r_d|\varphi_k, \sim) \propto \frac{1}{V^{|\mathbf{w}|}} \times \frac{1}{U} \times \frac{1}{S^{|\mathbf{w}|}} \times \sum_u \sum_\mathbf{s} \prod_{w \in \mathbf{w}} e^{-\frac{(r_d - \overline{r_{di}})^2}{2\sigma^2}}) \approx \frac{1}{V^{|\mathbf{w}|}} \times \frac{1}{U} \times \frac{1}{S^{|\mathbf{w}|}} \quad (10)$$

***Re-sample r (and s, u)***. In this step we re-sample word ratings $r_{di}$ of each word $w_{di}$.

$$p(r_{di} = r|\mathbf{r}^{-r_{di}}, \sim) = p(s_{di} = s, u_{di} = u|\mathbf{r}^{-r_{di}}, \sim)$$
$$\propto \frac{c_{kxu} + \eta}{c_{kx \cdot} + U\eta} \times \frac{l_{kws} + \lambda}{l_{kw \cdot} + S\lambda} \times e^{-\frac{(r_d - \overline{r_{di}})^2}{2\sigma^2}} \quad (11)$$

The Gibbs sampling runs for a sufficient number of iterations to repeatedly re-sample each conditional distribution of the state space $(\mathbf{t}, \mathbf{k}, \mathbf{r})$ according to equations (1) to (11). When all iterations are complete, we can estimate the following parameters.

$$\varphi_{kw} = \frac{l_{kw} + \beta}{l_{k \cdot} + V\beta}, \pi_{kws} = \frac{l_{kws} + \lambda}{l_{kw \cdot} + S\lambda}, \psi_{kxu} = \frac{c_{kxu} + \eta}{c_{kx \cdot} + U\eta} \quad (12)$$

### 3.3 Prediction

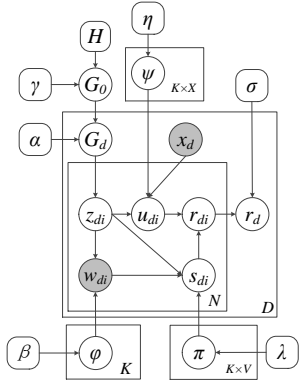

Fig 3 The graphic presentation of our prediction model

A regression model's performance is commonly evaluated by its ability to predict, so is our regression model. In prediction, our model is to estimate each rating $r_d$ of review $d$ by Gibbs sampling similar to what we described in section 3.2, however, this time $r_d$ is no longer observed, as shown as Fig 2, and it needs to be integrated out.

To integrate out $r_d$, equation (5) is modified as equation (13), where $\varphi_{kw}, \psi_{kxu}, \pi_{kws}$ come from the model trained in section 3.2 by equation (12).

$$f_k^{-w}(w|\varphi_k, \sim) \propto \int \sum_u \sum_s \frac{l_{kw} + \beta}{l_{k \cdot} + V\beta} \times \frac{c_{kxu} + \eta}{c_{kx \cdot} + U\eta} \times \frac{l_{kws} + \lambda}{l_{kw \cdot} + S\lambda} \times e^{-\frac{(r_d - \overline{r_{di}})^2}{2\sigma^2}} dr_d$$
$$\propto \int \sum_u \sum_s \varphi_{kw} \psi_{kxu} \pi_{kws} \times e^{-\frac{(r_d - \overline{r_{di}})^2}{2\sigma^2}} dr_d \quad (13)$$

In the special case when $k = k^{new}$, the equation is given by (14).

$$f_{k^{new}}^{-w}(w, r_d|\varphi_{k^{new}}, \sim) \propto \frac{1}{V} \times \frac{1}{U} \times \frac{1}{S} \times \int \sum_u \sum_s e^{-\frac{(r_d - \overline{r_{di}})^2}{2\sigma^2}} dr_d \quad (14)$$

Equation (9) is modified as

$$f_k^{-\mathbf{w}}(\mathbf{w}|\varphi_k, \sim) \approx f_k^{-\mathbf{w}}(\mathbf{w}, \mathbf{r}_d|\varphi_k, \sim) \propto (\prod_{w \in \mathbf{w}} \varphi_{kw}) \times \psi_{kxu} \times \prod_{w \in \mathbf{w}} \pi_{kws} \quad (15)$$

Equation (11) is modified as

$$p(r_{di} = r|\mathbf{r}^{-r_{di}}, \sim) \propto \psi_{kxu} \times \pi_{kws} \times \int e^{-\frac{(r_d - \overline{r_{di}})^2}{2\sigma^2}} dr_d \qquad (16)$$

Other equations of the prediction model's Gibbs sampling is the same as the regression model's Gibbs sampling described in section 3.2. When all iterations complete, we predict $r_d$ as the average of all non-neutral word ratings (17).

$$\hat{r}_d = \overline{r_{di}} \qquad (17)$$

## 4  Experiments & Evaluations

In this section, we describe the experiments and evaluate the rating prediction performance, effects of parameters, performance of sentiment analysis, and per-aspect sentiment/user preference. The Amazon datasets provided by https://snap.stanford.edu/data/web-Amazon.html are used to evaluate TSPRA's rating performance. Each dataset is a collection of reviews with each review associated with an author ID, a product ID and a rating. The rating is a 1-5 scale to indicate the customer's satisfactory level for the purchase. We put 80% of an author's reviews in the training set and the remaining reviews in the test set. To make user preference effective, we require that each author has at least 3 reviews in the training sets and 1 review in the test set. For datasets that have too many reviews meeting this requirement, we limit the training set to contain no more than 8000 reviews. The size of each training set and test set are shown in Table 2. The Amazon datasets provide realistic reviews for 22 different products, and we expect a comprehensive experiment on them would be convincing.

Table 2 Training sets and test sets

| Dataset | Training Set Size | Test Set Size | Dataset | Training Set Size | Test Set Size |
|---|---|---|---|---|---|
| Arts | 1960 | 894 | Jewelry | 7467 | 2338 |
| Auto | 7899 | 2187 | Musical Instrument | 5665 | 2510 |
| Baby | 932 | 538 | Office | 5901 | 2584 |
| Beauty | 7702 | 2569 | Patio | 3233 | 1092 |
| Cellphone | 3329 | 1857 | Shoes | 6967 | 2344 |
| Clothing | 7296 | 2385 | Software | 6075 | 2863 |
| Pet Supplies | 5743 | 2012 | Instant Video | 6090 | 2626 |
| Food | 6264 | 2309 | Tool & Home | 6529 | 2334 |
| Health | 3412 | 1144 | Toys | 5610 | 2737 |
| Kindle Store | 103 | 52 | Watches | 1126 | 636 |
| Industry & Science | 6353 | 2174 | Electronics | 5588 | 2671 |

Base on experiments of section 4.4 we choose $\mu = 3.5, \sigma^2 = 0.08$. The concentration parameters are set to $\gamma = 1.5, \alpha = 1.0, \beta = 0.5, \eta = 0.5, \lambda = 0.5$ following the default values in [4] [5]. Before the experiment, the review texts are pre-processed to remove punctuations, stop words and lemmatize using Stanford NLP [22]. The java implementation of our model and experiment results are currently accessible from https://github.com/tonyrivermsfly/TSPRA.

### 4.1  Prediction Performance Evaluation

For rating prediction we compare our model against a latest state-of-art model FLAME [9]. The performance measures we consider include

- Absolute error – the absolute value of true rating minus prediction, which is a very straightforward measure. The result is shown in Table 3, indicating our model reduces error in 18 of the 22 datasets, 9 of them achieves more than 10% reduction and with all datasets considered the average reduction is 6.7%.
- Pearson's correlation – measures how well the predicted ratings correlate with their corresponding true ratings. For two reviews with true ratings $r_1 > r_2$, a model with a higher correlation with the true values is in general less likely to give predictions $\hat{r}_1 < \hat{r}_2$. The results in Table 4 show our model performs well in this measure, and outperforms FLAME more than 15% on 10 of the 22 datasets.

- Number of inverted pairs – counts the actual number of mis-ordered pairs of predicted ratings, i.e. the number of pairs of reviews such that $r_1 > r_2$ but $\hat{r}_1 < \hat{r}_2$. The results in Table 5 shows our model achieves mild better performance in this measure.

We notice our model performs best in terms of error reduction on datasets cellphone (-16.5%), clothing (-14.6%), office (-13.4%) and pet (-12.0%). It has been discussed earlier in section 1 that decoupling sentiments and user preferences might bring potential performance improvement and make the model capable of discovering "critical aspects", those highly preferred aspects with relatively low sentiments. Later in section 4.3 we demonstrate the discovered obvious critical aspects in these four data sets.

Table 3 Rating Prediction Performance by Error
(lower values indicate better performance)

| Dataset | FLAME | TSPRA | | Dataset | FLAME | TSPRA | |
|---|---|---|---|---|---|---|---|
| Arts | 0.642 | 0.597 | -7.0% | Jewelry | 0.520 | 0.463 | -11.0% |
| Auto | 0.624 | 0.578 | -7.4% | Musical | 0.691 | 0.634 | -8.2% |
| Baby | 0.870 | 0.771 | -11.4% | Office | 0.650 | 0.563 | -13.4% |
| Beauty | 0.459 | 0.440 | -4.1% | Patio | 0.663 | 0.597 | -10.0% |
| Cellphone | 1.106 | 0.923 | -16.5% | Shoes | 0.346 | 0.365 | 5.5% |
| Clothing | 0.349 | 0.298 | -14.6% | Software | 1.001 | 0.899 | -10.2% |
| Pet | 0.807 | 0.710 | -12.0% | Video | 0.760 | 0.679 | -10.7% |
| Food | 0.832 | 0.863 | 3.7% | Tools | 0.735 | 0.665 | -9.5% |
| Health | 0.644 | 0.621 | -3.6% | Toys | 0.645 | 0.687 | 6.5% |
| Kindle | 1.119 | 1.197 | 7.0% | Watches | 1.017 | 0.950 | -6.6% |
| Industry | 0.370 | 0.345 | -6.8% | Electronics | 0.991 | 0.921 | -7.1% |

Table 4 Rating Prediction Performance by Pearson Correlation
(higher values indicate better performance)

| Dataset | FLAME | TSPRA | | Dataset | FLAME | TSPRA | |
|---|---|---|---|---|---|---|---|
| Arts | 0.716 | 0.768 | 7.3% | Jewelry | 0.707 | 0.772 | 9.2% |
| Auto | 0.697 | 0.755 | 8.3% | Musical | 0.534 | 0.621 | 16.3% |
| Baby | 0.501 | 0.631 | 25.9% | Office | 0.642 | 0.729 | 13.6% |
| Beauty | 0.816 | 0.834 | 2.2% | Patio | 0.742 | 0.792 | 6.7% |
| Cellphone | 0.405 | 0.619 | 52.8% | Shoes | 0.868 | 0.853 | -1.7% |
| Clothing | 0.897 | 0.926 | 3.2% | Software | 0.606 | 0.699 | 15.3% |
| Pet | 0.592 | 0.695 | 17.4% | Video | 0.627 | 0.729 | 16.3% |
| Food | 0.633 | 0.600 | -5.2% | Tools | 0.519 | 0.622 | 19.8% |
| Health | 0.735 | 0.758 | 3.1% | Toys | 0.665 | 0.601 | -9.6% |
| Kindle | 0.780 | 0.736 | -5.6% | Watches | 0.430 | 0.518 | 20.5% |
| Industry | 0.848 | 0.870 | 2.6% | Electronics | 0.503 | 0.587 | 16.7% |

Table 5 Rating Prediction Performance by Inverted Pairs
(lower values indicate better performance)

| Dataset | FLAME | TSPRA | | Dataset | FLAME | TSPRA | |
|---|---|---|---|---|---|---|---|
| Arts | 495 | 481 | -2.8% | Jewelry | 2326 | 2275 | -2.2% |
| Auto | 2362 | 2273 | -3.8% | Musical | 1468 | 1451 | -1.2% |
| Baby | 338 | 325 | -3.8% | Office | 1834 | 1769 | -3.5% |
| Beauty | 1757 | 1711 | -2.6% | Patio | 805 | 829 | 3.0% |
| Cellphone | 2196 | 2083 | -5.1% | Shoes | 1745 | 1851 | 6.1% |
| Clothing | 1550 | 1366 | -8.8% | Software | 4108 | 3994 | -2.8% |
| Pet | 1799 | 1749 | -2.8% | Video | 2764 | 2716 | -1.7% |
| Food | 1727 | 1756 | 1.7% | Tools | 1753 | 1735 | -1.0% |
| Health | 745 | 743 | -0.3% | Toys | 1960 | 2083 | 6.3% |
| Kindle | 32 | 36 | 12.5% | Watches | 537 | 510 | -5.0% |
| Industry | 1072 | 1010 | -5.8% | Electronics | 2402 | 2391 | -0.5% |

## 4.2 Sentiment Analysis

Our model is not only a rating regression model, but also a sentiment analysis model that can automatically identify continuous word sentiment polarity values from the product reviews. In the real world most product reviews are typically annotated by ratings, thus review rating prediction might not be of much practical use, although it serves as an important indicator of model performance as shown in section 4.1. On the contrary, word sentiments derived from the regression process could be much more valuable to business. Although many previous papers [2, 3, 8, 9, 15-17] include word sentiments as model factors, none of them demonstrate the sentiment analysis results in detail, especially the polarity evaluations.

In this section we compare the our model's sentiment polarity evaluations to SenticNet, which is first introduced in [23]. It is a public sentiment resource built by non-probabilistic approaches, where each word is also associated with a polarity value ranging from -1 to 1 to quantify from the most negative sentiment to the most positive sentiment. The latest version is SenticNet3 [24]. Due to the unavailability of sentiment analysis results of previous review models, we opt to compare our results with this public sentiment resource.

As explained in section 3.1 and 3.2, our model infers each word's sentiment distribution and stores them in matrix $\pi_{K \times V}$. Recall that an element $\pi_{zw}$ of this matrix is a three dimensional categorical distribution, indicating the probability of word $w$ being positive (let's denote it as $\pi_{zw}^+$), negative (denoted as $\pi_{zw}^-$) or neutral (denoted as $\pi_{zw}^0$) under topic $z$. As before the noisy neutral probability is ignored and the polarity value of word $w$ is derive by formula (18).

$$\text{polarity}(w) = \frac{\sum_{z=1}^{K}(\pi_{zw}^+ - \pi_{zw}^-)}{\sum_{z=1}^{K}(\pi_{zw}^+ + \pi_{zw}^-)} \tag{18}$$

The vocabulary of the review texts has a 2524-word intersection with SenticNet3. The polarity distributions of this intersection are shown in Fig 4. From Fig 4 we see most of Sentic3 polarities are squeezed around the neutral polarity 0, while polarity estimations by our model TSPRA are more evenly distributed. Our estimations are right-skewed, which is caused by the unbalanced amount of 5-star review ratings in the datasets, see Table 2. In the future we might want to perform a balanced sampling from the datasets and re-run the experiments.

We demonstrate the top 20 positive words and top 20 negative word identified by our model in

Table 6 with comparison of their polarity values to SenticNet3 values, and our estimations are clearly a better match to our intuition in the context of review ratings. For example, "delicious", "tasty", "comfy" are expressing very positive sentiments in the context of product reviews (mostly food, we think), for which SenticNet3 only assigns quite unreasonable near-zero neutral values 0.05, 0.11, and 0.06. For another example, SenticNet3 assigns the obvious negative word "unreliable" and "awful" positive values 0.12, 0.12, while by contrast our model assigns -0.88, -0.78. The experiment of this section confirms that our model is also a good sentiment analysis model in addition to its ability of rating prediction.

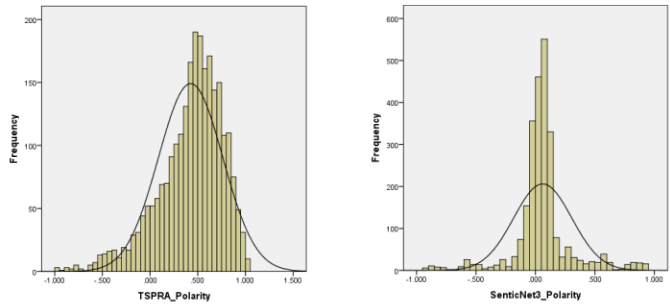

Fig 4 Polarity distributions of our model and SenticNet3.

Table 6 Polarity value comparison between our model and SenticNet3

|  | TSPRA | SNet3 |  | TSPRA | SNet3 |
| --- | --- | --- | --- | --- | --- |
| great | 0.99 | 0.86 | dreadful | -1.00 | -0.29 |
| love | 0.99 | 0.32 | rude | -1.00 | -0.84 |
| tasty | 0.99 | 0.05 | piss | -0.95 | -0.04 |
| perfect | 0.98 | 0.58 | supervisor | -0.90 | 0.04 |
| best | 0.98 | 0.22 | unreliable | -0.88 | 0.13 |
| delicious | 0.97 | 0.11 | worst | -0.86 | -0.12 |
| excellent | 0.97 | 0.50 | disappointment | -0.84 | -0.06 |
| awesome | 0.97 | 0.78 | disappointed | -0.80 | -0.17 |
| easy | 0.96 | 0.27 | waste | -0.78 | -0.06 |
| beautiful | 0.95 | 0.15 | awful | -0.78 | 0.12 |
| well | 0.95 | 0.72 | poorly | -0.77 | -0.07 |
| nice | 0.95 | 0.16 | terrible | -0.76 | -0.90 |
| yummy | 0.94 | 0.06 | useless | -0.74 | -0.26 |
| pleased | 0.94 | 0.45 | return | -0.64 | 0.14 |
| amazing | 0.94 | 0.36 | nightmare | -0.62 | -0.10 |
| comfort | 0.94 | 0.33 | horrible | -0.60 | -0.94 |
| wonderful | 0.94 | 0.81 | disgust | -0.59 | -0.29 |
| classy | 0.93 | 0.10 | overprice | -0.56 | 0.04 |
| comfy | 0.93 | 0.15 | error | -0.56 | -0.27 |
| happy | 0.92 | 0.30 | worse | -0.56 | 0.13 |

Top 20 positive words and top 20 negative words identified by our model are chosen for comparison with generally non-sentiment nouns like "mustang, doll, film" removed. "great" is the most positive word recognized by our model, and "dreadful" is the most negative word identified by our model. Those who are interested in the full list can view our public sentiment file https://github.com/tonyrivermsfly/TSPRA/blob/master/experiments/Word%20Sentiments.xlsx.

### 4.3 Critical Aspects

Due to the decoupling of user preference from sentiment, we find another practical use of our model in addition to sentiment analysis -- identifying those product aspects with high user preference but low sentiments (aspect sentiment is derived from word sentiments). The results can help business decision makers to quickly recognize certain important product aspects they should pay attention to, which we call "*critical aspects*".

In our model TSPRA, the $K \times X$ matrix $\psi$ stores the inferred users' preferences distribution over product aspects. Recall that an element $\psi_{zx}$ of this matrix is a 0-1 distribution, indicating the probability of user $x$ (let's denote it as $\psi_{zx}^+$) having strong preference to product aspect $z$, and the probability of weak preference to product aspect $z$. Thus the average aspect preference of a product aspect $z$ can be calculated by formula (19), and its value ranges from 0 to 1.

$$\text{preference}(z) = \frac{\sum_{x=1}^{X} \psi_{zx}^+}{X} \tag{19}$$

Similar to formula (18) in section 4.2, formula (20) is used to calculate aspect sentiment and its value ranges from -1 to 1.

$$\text{sentiment}(z) = \frac{\sum_{w=1}^{V}(\pi_{zw}^+ - \pi_{zw}^-)}{\sum_{w=1}^{V}(\pi_{zw}^+ + \pi_{zw}^-)} \tag{20}$$

We then identify those aspects with minimum preference 0.3 and preference-sentiment ration being negative or larger than 2. These aspects are concerned by users yet quite negatively or less positively commented. We define these aspects as "critical aspects", meaning that business decision makers should really pay attention to these aspects, and improvement to these aspects could be most effective to enhance user experience.

The results for three datasets "cellphone", "clothing" and "office" are shown in Table 7, where 4 critical aspects can be observed. Most frequent words of the 5 critical aspects are shown in Table

8. Based on both tables we can clearly see "battery" and "phone call service" are aspects of "cellphone" that are of relatively high user concern but plagued by negative sentiment. Aspect "jeans" of product "clothing", and aspect "phone" of product "office" are also such critical aspects.

Table 7 Demonstration of critical aspects

|   | Cellphone | | Clothing | | Office | | Pet | |
|---|---|---|---|---|---|---|---|---|
|   | Pref | Senti | Pref | Senti | Pref | Senti | Pref | Senti |
| 1 | 0.650 | 0.606 | 1.000 | 0.629 | 1.000 | 0.767 | 0.791 | 0.695 |
| 2 | 1.000 | 0.526 | 0.037 | 0.653 | **0.439** | **0.164** | 0.688 | 0.525 |
| 3 | 0.408 | 0.389 | **0.352** | **0.058** | 0.155 | 0.372 | 0.428 | 0.512 |
| 4 | **0.556** | **0.145** | 0.204 | 0.643 | 0.403 | 0.697 | 0.236 | 0.514 |
| 5 | **0.447** | **-0.539** | 0.709 | 0.714 | 0.027 | 0.058 | 0.460 | 0.534 |
| 6 | **0.951** | **0.308** | 0.160 | 0.646 | 0.307 | 0.637 | 0.001 | -0.463 |
| 7 | 0.008 | 0.244 | | | 0.200 | 0.446 | **1.000** | **0.402** |
| 8 | 0.177 | 0.252 | | | | | 0.010 | 0.202 |
| 9 | 0.019 | 0.156 | | | | | 0.004 | 0.328 |

Table 8 Most frequent words of the critical aspects in Table 7

| Cellphone Topic 4 (battery) | battery, charge, charger, buy, product, motorola, price, cable, usb, plug |
|---|---|
| Cellphone Topic 5 (call service) | service, call, minute, nextel, cingular, customer, try, free, number, plan |
| Clothing Topic 3 (jeans) | jeans, size, fit, order, pair, look, tight, leg, waist, skinny |
| Office Topic 2 (phone) | phone, handset, call, base, number, feature, quality, sound, system, cordless |
| Pet Topic 7 (animal toys) | toy, cat, dog, bird, like, chew, litter, ball, box, buy, play, time, puppy |

At the end we examine the Pearson's correlation between aspect preferences and aspect sentiments, which is a weak value 0.349 and it confirms that our decouple of these two types of variables in our model is valid. As mentioned in introduction, a distinction between TSPRA and collaborative filtering models like [8, 9] is that we define "preference" as how much customers care about a product aspect and by definition it is independent from word sentiments. In CF models "preference" is viewed more like aspect-level sentiment that causes word or sentence sentiments, which must have a strong correlation with the aggregation of word or sentence sentiments. The unbalanced amount of positive ratings, which possibly contributes to the right-skewness of the word sentiment polarity distribution, could also be the reason of the remaining correlation.

### 4.4 Experiments on Parameters

The neutral rating $\mu$ and rating noise $\sigma$ are two main parameters introduced by our model TSPRA. In this section we experiment on the effects of both parameters in terms of error. We choose 6 datasets with different average review ratings to test the parameter $\mu$ by fixing $\sigma^2 = 0.08$, and then we test $\sigma^2$ by fixing $\mu = 3.5$ on the same 6 datasets. The results are shown in Fig 5.

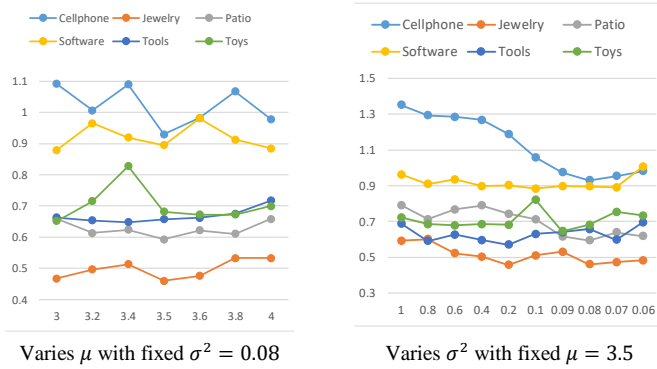

| Varies $\mu$ with fixed $\sigma^2 = 0.08$ | Varies $\sigma^2$ with fixed $\mu = 3.5$ |

Fig 5 Error varies with neutral rating parameter **$\mu$** and noise parameter **$\sigma^2$**.

As we can see there is no strong agreement on the best parameter among the datasets. That is not surprising since it is reasonable that different products have somewhat different neutral rating and rating noise. However, we do observe from the above charts that the model archives relatively good results on all six data sets when $\mu$ is around 3.5 and $\sigma^2$ is around 0.08. Hence we recommend $\mu = 3.5, \sigma^2 = 0.08$ as the default values, as what we have done in section 4.1.

Also we find $\sigma^2 = 0.08$ is actually making sense. The parameter value of $\sigma^2$ reflects our belief how the majority of users give the final rating based on the average of non-neutral word ratings (the neutral words are ignored when review rating is estimated, see section 3.1 and 3.2). Under normal distribution, 68% of final ratings vary from $(\overline{r_{di}} - \sigma, \overline{r_{di}} + \sigma)$ and 95% of final ratings vary from $(\overline{r_{di}} - 2\sigma, \overline{r_{di}} + 2\sigma)$. When we set $\sigma^2 = 0.08$ and thus $\sigma \approx 0.283$, it actually indicates 68% of review ratings are in the range of $(\overline{r_{di}} - 0.283, \overline{r_{di}} + 0.283)$ and 95% of review ratings are in the range of $(\overline{r_{di}} - 0.566, \overline{r_{di}} + 0.566)$. In plain words this implies users are rounding their review ratings. For example, if a user wants to rate a product as 3.7, then due to limit of the 1-5 scale, the user has to round the rating to 4.0.

## 5  Conclusion

We have proposed a model that considers topics, sentiments and user preferences as integrated factors to explain review texts and ratings. Two distinctions between our model and previous models are the de-correlation between preference and sentiment, and the adoption of HDP framework that automatically identifies the topic number of each dataset. Our model outperforms the FLAME model proposed by [9] in terms of error, Pearson's correlation and number of inverted pairs. Further experiment on sentiment analysis show our model can yield very reasonable word sentiments, most of them make more sense in comparison to a public sentiment resources SenticNet3 [24] constructed via non-probabilistic approaches. A third experiment shows our model is able to capture the "critical aspects" – the negatively commented product aspects concerned a lot by users. Improving the critical aspects would be most effective to enhance overall user experience. Last but not the least, the correlation between "preference" and "sentiment" in our model is weak, verifying our claim that these two concepts are largely independent factors.